\documentclass[11pt,draftcls,onecolumn] {IEEEtran}
\usepackage[mathscr]{eucal}
\usepackage[cmex10]{amsmath}
\usepackage{epsfig,epsf,psfrag}
\usepackage{amssymb,amsmath,amsthm,amsfonts,latexsym}
\usepackage{amsmath,graphicx,bm,xcolor,url}
\usepackage[caption=false]{subfig} 
\usepackage{fixltx2e}%ordering of single and double column floats
\usepackage{array}%array and tabular environments
\usepackage{verbatim}
\usepackage{bm}
\usepackage{algorithmic, cite}
\usepackage{algorithm}
\usepackage{verbatim}
\usepackage{textcomp}
\usepackage{mathrsfs}
\usepackage{epstopdf}
\catcode`~=11 \def\UrlSpecials{\do\~{\kern -.15em\lower .7ex\hbox{~}\kern .04em}} \catcode`~=13 

\allowdisplaybreaks[3]

\newcommand{\calI}{\mathcal{I}}

\newcommand{\calN}{\mathcal{N}}
\newcommand{\calO}{\mathcal{O}}

\newcommand{\calR}{\mathcal{R}}
\newcommand{\calS}{\mathcal{S}}

\newcommand{\ba}{\mathbf{a}}
\newcommand{\bA}{\mathbf{A}}

\newcommand{\bC}{\mathbf{C}}
\newcommand{\bd}{\mathbf{d}}
\newcommand{\bD}{\mathbf{D}}

\newcommand{\bh}{\mathbf{h}}

\newcommand{\bq}{\mathbf{q}}
\newcommand{\bQ}{\mathbf{Q}}

\newcommand{\bS}{\mathbf{S}}

\newcommand{\bT}{\mathbf{T}}

\newcommand{\bU}{\mathbf{U}}

\newcommand{\bx}{\mathbf{x}}
\newcommand{\bX}{\mathbf{X}}
\newcommand{\by}{\mathbf{y}}
\newcommand{\bY}{\mathbf{Y}}

\DeclareMathAlphabet{\mathbsf}{OT1}{cmss}{bx}{n}
\DeclareMathAlphabet{\mathssf}{OT1}{cmss}{m}{sl}% slanted sans serif

\DeclareSymbolFont{bsfletters}{OT1}{cmss}{bx}{n}  
\DeclareSymbolFont{ssfletters}{OT1}{cmss}{m}{n}
\DeclareMathSymbol{\bsfGamma}{0}{bsfletters}{'000}
\DeclareMathSymbol{\ssfGamma}{0}{ssfletters}{'000}
\DeclareMathSymbol{\bsfDelta}{0}{bsfletters}{'001}
\DeclareMathSymbol{\ssfDelta}{0}{ssfletters}{'001}
\DeclareMathSymbol{\bsfTheta}{0}{bsfletters}{'002}
\DeclareMathSymbol{\ssfTheta}{0}{ssfletters}{'002}
\DeclareMathSymbol{\bsfLambda}{0}{bsfletters}{'003}
\DeclareMathSymbol{\ssfLambda}{0}{ssfletters}{'003}
\DeclareMathSymbol{\bsfXi}{0}{bsfletters}{'004}
\DeclareMathSymbol{\ssfXi}{0}{ssfletters}{'004}
\DeclareMathSymbol{\bsfPi}{0}{bsfletters}{'005}
\DeclareMathSymbol{\ssfPi}{0}{ssfletters}{'005}
\DeclareMathSymbol{\bsfSigma}{0}{bsfletters}{'006}
\DeclareMathSymbol{\ssfSigma}{0}{ssfletters}{'006}
\DeclareMathSymbol{\bsfUpsilon}{0}{bsfletters}{'007}
\DeclareMathSymbol{\ssfUpsilon}{0}{ssfletters}{'007}
\DeclareMathSymbol{\bsfPhi}{0}{bsfletters}{'010}
\DeclareMathSymbol{\ssfPhi}{0}{ssfletters}{'010}
\DeclareMathSymbol{\bsfPsi}{0}{bsfletters}{'011}
\DeclareMathSymbol{\ssfPsi}{0}{ssfletters}{'011}
\DeclareMathSymbol{\bsfOmega}{0}{bsfletters}{'012}
\DeclareMathSymbol{\ssfOmega}{0}{ssfletters}{'012}

\DeclareMathOperator*{\argmin}{arg\,min}

\newtheorem{theorem}{Theorem} 
\newtheorem{lemma}[theorem]{Lemma}

\newtheorem{definition}{Definition}

\newtheorem{remark}{Remark}

\newtheorem{assumption}{Assumption}
\newtheorem{data model}{Data Model}

\newcommand{\qednew}{\nobreak \ifvmode \relax \else
      \ifdim\lastskip<1.5em \hskip-\lastskip
      \hskip1.5em plus0em minus0.5em \fi \nobreak
      \vrule height0.75em width0.5em depth0.25em\fi}

\usepackage{makecell}
\begin{document}
% \title{\huge{Spatially Random Column/Feature Sampling: A Simple, Randomized, and Structure-Preserving Data Sketching Tool}}
\title{\huge{Spatial Random Sampling: A Structure-Preserving Data Sketching Tool}}

\author{Mostafa~Rahmani, \IEEEmembership{Student Member,~IEEE} and George~K.~Atia,~\IEEEmembership{Member,~IEEE} % <-this % stops a space
\thanks{
\noindent
This work is supported in part by
NSF CAREER Award CCF-1552497 and NSF Grant CCF-1320547.

The authors are with the Department of Electrical and Computer Engineering, University of Central Florida, Orlando,
FL 32816 USA (e-mail: mostafa@knights.ucf.edu, george.atia@ucf.edu).}% <-this % stops a spa
}

\markboth{}%
%\markboth{Journal of \LaTeX\ Class Files,~Vol.~11, No.~4, %December~2012}%
{Shell \MakeLowercase{\textit{et al.}}: Bare Demo of IEEEtran.cls for Journals}
\maketitle


\begin{abstract}
% Whereas random column sampling is largely utilized to obtain small representative data sketches, it is not guaranteed to preserve the underlying structure of the data and is prone to missing data clusters with smaller population sizes.
Random column sampling is not guaranteed to yield data sketches that preserve the underlying structures of the data and may not sample sufficiently from less-populated data clusters.
Also, adaptive sampling can often provide accurate low rank approximations, yet may fall short of producing descriptive data sketches, especially when the cluster centers are linearly dependent.
Motivated by that, this paper introduces a novel randomized column sampling tool dubbed Spatial Random Sampling (SRS), in which data points are sampled based on their proximity to randomly sampled points on the unit sphere.
% the magnitude of data projections along a set of random directions.
% The normalized data is first projected along a set of random directions to select the data points to sample.
The most compelling feature of SRS is that the corresponding probability of sampling from a given data cluster is proportional to the surface area the cluster occupies on the unit sphere, independently of the size of the cluster population.
% occupied by that data cluster on the unit sphere and is independent from the cluster population.
Although it is fully randomized, SRS is shown to provide descriptive and balanced data representations. The proposed idea addresses a pressing need in data science and holds potential to inspire many novel approaches for analysis of big data.
\end{abstract}


\section{Introduction}
The complexity of many of the existing data analysis and machine learning algorithms limits their scalability to high-dimensional settings.
% The complexity of most of data analysis and machine learning algorithm are not scalable to high dimensional applications.
This has spurred great interest in data sketching techniques that produce descriptive and representative sketches of the data on the premise that substantial complexity reductions can be potentially achieved without sacrificing performance when data inferencing is carried out using such sketches in lieu of the full-scale data
\cite{mahoney2009cur,woolfe2008fast,sun2007less,martinsson2010normalized,affandi2013nystrom,rahmani2016high,halko2011algorithm,rahmani2016randomized,boutsidis2010random,farahat2015greedy,boutsidis2009unsupervised,rahimi2007random,ailon2006approximate}. 
 % , yielding notable speedups without sacrificing performance.
 % Data sketches are a set of linear or non-linear observations of the data which can reveal or represent the important information and the structures of the big data. The basic idea for big data analysis is to utilize the data sketches as apposed to the whole high dimensional data.

Random embedding and random column sampling are two widely used linear data sketching tools. Random embedding projects data in a high-dimensional space onto a random low-dimensional subspace, and was shown to notably reduce dimensionality while preserving the pairwise distances between the data points \cite{ailon2006approximate,achlioptas2003database,johnson1984extensions}. While random embedding can generally preserve the structure of the data, it is not suitable for feature/column sampling. In random column sampling, a column is selected via random sampling from the column index set -- hence the alternative designation Random Index Sampling (RIS). For RIS, the probability of sampling from a data cluster is clearly proportional to its population size. As a result, RIS may fall short of preserving structure if the data is unbalanced, in the sense that RIS may not sample sufficiently from less-populated data clusters, and/or may not capture worthwhile features that could be pertinent to rare events. This motivates the work of this paper in which we develop a new random sampling tool that can yield a descriptive data sketch even if the given data is largely unbalanced.

Over the last two decades, many different column sampling methods were proposed \cite{lamport14}.
Most of these methods aim to find a small set of informative data columns whose span can well approximate the given data. In other words, if $\bC \in \mathbb{R}^{N_1 \times n}$ is the matrix of sampled columns, where $n$ is the number of sampled columns and $N_1$ the ambient dimension, most of the existing column sampling methods seek a solution to the optimization problem
\begin{eqnarray}
\min_{\bC} \:\: \| \bD - \bC \bC^{\dagger} \bD \|_F,
\label{eq:basic}
\end{eqnarray}
where $\bD \in \mathbb{R}^{N_1 \times N_2}$ denotes the data, $^\dagger$ the pseudoinverse, and $\| 	\cdot \|_{F} $  the Frobenius norm. These methods
 can be broadly categorized into randomized \cite{deshpande2006matrix,nguyen2009fast,boutsidis2009improved,rudelson2007sampling,guruswami2012optimal,paul2015column,drineas2004clustering,drineas2006subspace} and deterministic methods
 \cite{rahmani2016robust,balzano2010column,farahat2013efficient,civril2012column,boutsidis2008clustered,lashkari2007convex,esser2012convex,elhamifar2012see,gu1996efficient}. In randomized methods, the columns are sampled based on a carefully chosen probability distribution. For instance, \cite{drineas2004clustering} uses the $\ell_2$-norm of the columns,
and in \cite{drineas2006subspace} the sampling probabilities are proportional to
the norms of the rows of the top right singular vectors. There are different types of deterministic sampling
algorithms, including the rank revealing QR algorithm \cite{gu1996efficient}, and clustering-based algorithms \cite{boutsidis2008clustered}. In \cite{elhamifar2012see,misra2014data,nie2010efficient,rahmani2016robust}, the non-convex optimization (\ref{eq:basic}) is relaxed to a convex program by finding a row-sparse representation of the data.
We refer the reader to
\cite{lamport14,new4,elhamifar2012finding} and references therein for more information about the matrix approximation based sampling methods.

Whereas low rank approximation has been instrumental in many applications, the sampling algorithms based on (\ref{eq:basic}) cannot always guarantee that the sampled points satisfactorily capture the structure of the data. For instance, suppose the columns of $\bD$ form $m$ clusters in the $N_1$-dimensional space, but the cluster centers are  linearly dependent.  An algorithm which aims to minimize (\ref{eq:basic}) would not necessarily sample from each data cluster since it only looks for a set of columns whose span is that of the dominant singular vectors of $\bD$.

%\smallbreak
% \noindent
% \textbf{Notation:}
For notation, bold-face upper-case letters denote matrices and bold-face lower-case letters denote vectors. Given a matrix $\bA$, $\ba_i$ and $\ba^i$ denote its $i^\text{th}$ column and $i^\text{th}$ row, respectively. For a vector $\ba$, $\max \ba$ is its maximum element, $|\ba|$ the vector of absolute values of its elements, and $\ba_{\cal I}$ for an index set $\cal I$ the elements of $\ba$ indexed by $\cal I$. 
Also, $\mathbb{S}^{N_1-1}$ designates the unit $\ell_2$-norm sphere in $\mathbb{R}^{N_1}$. 

\section{Proposed Method}
% In many applications (e.g. clustering and classification), the data sketch should preserve the structure of data.
% An important task of data sketching tools is to produce structure-preserving data sketches.
%
As mentioned earlier, with RIS the probability of sampling from a data cluster is proportional to its population size. However, in many applications of interest the desideratum is to collect more samples from clusters that occupy a larger space
or that have higher dimensions, thereby composing a structure-preserving sketch of the data.
% a sampling probability that is proportional to the \textcolor{red}{volume of the} space occupied by that cluster.
% it is a more efficient policy to have a sampling algorithm where the probability of sampling from a data cluster is proportional to the space occupied by the data cluster.
For instance, suppose the data points lie on $\mathbb{S}^{N_1 -1}$ and form two linearly separable clusters such that the surface area corresponding to the first cluster on the unit sphere is greater than that of the second cluster\footnote{The notion of surface area for comparing the spatial distribution of clusters will be made precise in Definition \ref{def:surf_area} in the next subsection.}. In this case, a structure-preserving sketch should generally comprise more data points from the first cluster. However, RIS would sample more points from the cluster with the larger population regardless of the structure of the data.

\subsection{Spatially random column sampling}
When the data points are projected onto $\mathbb{S}^{N_1 - 1}$, each cluster will occupy a certain surface area on the unit sphere.
% assume that the projection of data clusters on  the unit sphere    are separable.
We propose a random column sampling approach in which the probability of sampling from a data cluster is proportional to its corresponding surface area. 
%In this paper, we focus on the case where the surface areas corresponding to  different clusters do not overlap. 
The proposed method, dubbed Spatially Random column Sampling (SRS), is presented in the table of Algorithm 1 along with the definitions of the used symbols. SRS samples the $n$ data points whose normalized versions have the largest projections along $n$ randomly selected directions in $\mathbb{R}^{N_1}$ (the rows of matrix $\mathbf{\Phi}$). Unlike RIS, SRS performs random sampling in the spatial domain as opposed to the index domain, wherefore the probability of  sampling from a data cluster depends on its spatial distribution. 
%\textcolor{blue}{Is it really independent? For example, since sampling is without replacement it is certainly different to have a cluster with two points versus one with 1000 points. We need to be absolutely accurate.}
%
%With the Random Index Sampling (RIS), each column index has an equal chance to be sampled by the sampling algorithm. Accordingly, the probability of sampling from each data cluster is proportional to the population of the data cluster.
%In many applications (e.g. clustering and classification), we  want to preserve the structure of the data distribution in the space. Therefore,
%in many cases the objective is to have a sampling algorithm such that the probability of sampling from each data cluster is proportional to the space occupied by the data cluster. This is a more reasonable sampling policy for many machine learning application since for a data cluster with a larger space, we need more data points to  represent it. For instance, the number of data points one needs to represent a subspace is proportional to the dimensional of the subspace. Therefore, we propose
%Spatially Random column Sampling (SRS) method (Algorithm 1) which performs the random sampling in the space domain as opposed to the index domain, i.e.,  the random sampling is carried out on the unit $\ell_2$-norm sphere as apposed to the column indexes.
%
%\smallbreak
% \noindent
%\emph{Insight:}
To provide some insight into the operation of SRS, consider the following fact \cite{cai2013distributions,cai2012phase}.
% $\mathbf{\phi}^i / \| \mathbf{\phi}^i \|_2$, where $\mathbf{\phi}^i$ is the $i^{\text{th}}$ row of $\mathbf{\Phi}$, has a uniform distribution on the unit sphere $\mathbb{S}^{N_1 -1}$.
\begin{lemma}
If the elements of $\bm{\phi} \in \mathbb{R}^{N_1}$ are sampled independently from $\calN (0,1)$, the vector $\bm{\phi} / \| \bm{\phi} \|_2$
will have a uniform distribution on the unit $\ell_2$-norm sphere $\mathbb{S}^{N_1 - 1}$.
\label{lm:random vectors}
\end{lemma}

\noindent
According to Lemma \ref{lm:random vectors}, $\bm{\phi}^i / \| \bm{\phi}^i \|_2$ % each row of $\mathbf{\Phi}$
corresponds to a random point on the unit sphere (recalling that $\bm{\phi}^i$ is the $i^{\text{th}}$ row of $\mathbf{\Phi}$).
The probability that a random direction lies in a given cluster is proportional to its corresponding surface area on the unit sphere. Since we cannot ensure if a random direction lies in a data cluster, we sample the data point at minimum distance from the randomly sampled direction. Therefore, Algorithm 1 samples $n$ points randomly on the unit sphere, and for each randomly sampled direction samples the data points with closest proximity to that direction. As such, it is more likely to sample from a data cluster that covers a larger area on the unit sphere. %  it has a higher chance to be sampled from.
More precisely, suppose the columns of $\bD$ have unit $\ell_2$-norm and 
form $s$ separable clusters. We divide the surface area of the unit sphere into $s$ regions $\{ \calR_i \}_{i=1}^s$, where $\calR_i$ is defined as follows. 
\begin{definition}
\label{def:surf_area}
Suppose the matrix $\bD$ can be represented as $\bD = [\bD_1 \: ... \: \bD_s]\bT$, where  $\bD_k $ consists of the data points in the $k^{\text{th}}$ cluster, $\bT$ is a permutation matrix, and the columns of $\bD$ have unit $\ell_2$-norm. The region $\calR_i$ is defined as
\[
\calR_i := 
\left\{ \by \in \mathbb{S}^{N_1-1} \Big| \max |\by^T \bD_i| > \max |\by^T \bD_j|, \: \forall\: j \neq i \right\}.
\]
\end{definition} 

\noindent 
Accordingly, the probability of sampling a data point from the $k^{\text{th}}$ cluster is linear in the area of $\calR_k$.
\begin{remark}
Define $\bU$ as an orthonormal basis for the column space of $\bD$. Since the rows of $\mathbf{\Phi}$ are random vectors, $\mathbb{P} [ \bU^T \bm{\phi}^i  = 0] = 0, \forall i$, where $\mathbb{P}$ denotes the probability measure. 
In addition, since $\bm{\phi}^i / \| \bm{\phi}^i\|_2$ has a uniform distribution on $\mathbb{S}^{N_1 -1}$, $\bU \bU^T \bm{\phi}^i / \| \bU^T \bm{\phi}^i \|_2$ has a uniform distribution on the intersection of the column space of $\bU$ and $\mathbb{S}^{N_1 -1}$. Thus, SRS generates $n$ random directions in the span of the data. 
%\textcolor{blue}{What do you mean by are equivalent? Also, what is the use of this remark?}
\end{remark}

%\begin{remark}
%In Algorithm 1, the projection of the data points on different random directions is performed sequentially. Algorithm 2 provides an alternative batch implementation of the second step of Algorithm 1 where all the projections are carried out using one matrix multiplication. It leads to a notably higher running time when $n$ is large. \textcolor{blue}{Higher running time? Also, do we really need a table for Algorithm 2. Can't we just say it in words? This is just a simple modification where all the inner products with all directions are computed at once.}
%\end{remark}
%
In the following section, we compare the requirements of RIS and SRS (as two completely random column sampling tools) using a set of theoretical examples.

\begin{algorithm}
\caption{Spatially random column sampling (SRS without replacement)}
{\footnotesize
\textbf{Input}: Data matrix $\bD \in \mathbb{R}^{N_1 \times N_2} $ and $n$ as the number of sampled columns.

\smallbreak
\textbf{Initialization}: Construct matrix $\mathbf{\Phi} \in \mathbb{R}^{n \times N_1}$ by sampling independently from $\calN (0,1)$. Set $\bY$ equal to an empty matrix.

\smallbreak
\textbf{1. Data Normalization}:
Define $\bX \in \mathbb{R}^{N_1 \times N_2}$ such that $\bx_i = \bd_i / \| \bd_i \|_2$.

\smallbreak
\textbf{2. Column Sampling}:\\
\textbf{2.1} Set $\bQ =    \mathbf{\Phi} \bX  $ and set $\calI = \emptyset$.

\textbf{2.2 For} $i = 1$ to $n$\\
\textbf{2.2.1} Define $\bh = \left| \bq^i \right|$ and set $\bh_{\calI} = 0$. \\
\textbf{2.2.2} Define $\bx_k$ as the $k^{\text{th}}$ column of $\bX$, where $k$ is the index of the maximum element of $\bh$. \\
\textbf{2.2.3} Update $\bY = [\bY \:\: \bx_k]$ and add $k$ to the set $\calI$.\\
\textbf{2.2 End For}

\smallbreak
\textbf{Output:}  $\bY \in \mathbb{R}^{N_1 \times n}$ is the matrix of sampled columns.

 }
\end{algorithm}
\vspace{-.35cm}

\begin{algorithm}
\caption{SRS with replacement}
{\footnotesize

\textbf{1.} Perform step 1 of Algorithm 1 and set $\bQ =    \mathbf{\Phi} \bX  $.

\smallbreak
\textbf{2. Column Sampling:} Matrix $\bY \in \mathbb{R}^{N_1 \times n}$ is the matrix of sampled columns. The $i^{\text{th}}$ column of $\bY$, $\by_i$, is equal to $\bx_k$ where $k$ is equal to the index of the maximum element of $| \bq^i |$.

}
\end{algorithm}

%\subsection{Theoretical examples}
\subsection{Sample complexity analysis}
This section provides a theoretical analysis of the sample complexity of SRS in the context of two examples, in which we show that the probability of sampling from a data cluster with SRS can be independent of the cluster population. The sample complexity is contrasted to that of conventional RIS. To simplify the analysis, we assume that sampling in SRS is done with replacement (c.f. Algorithm 2).
\begin{figure}[t!]
    \includegraphics[width=0.7\textwidth]{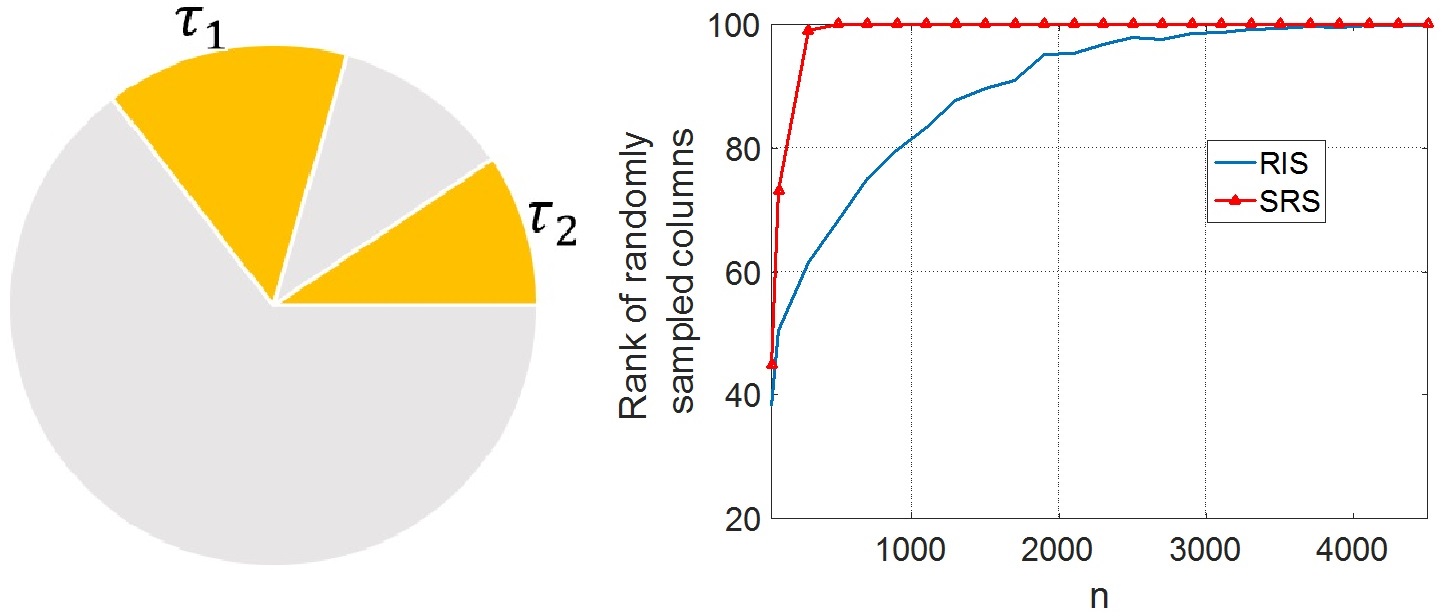}
    \centering
     \vspace{-0.1in}
    \caption{Left: The distribution of data in a two-dimensional space. The data consists of two clusters shown in yellow. The normalized points in the first and second clusters are distributed on two separate arcs with lengths $\tau_1$ and $\tau_2$, respectively. Each of the two arcs does not overlap with the image of the other arc w.r.t. the origin on the unit circle. Right: The rank of randomly sampled columns by RIS and SRS versus the number of sampled columns.}
 %   \label{fig: phases}
%   \vspace{-.5cm}
\label{fig:circlerank}
\end{figure}

\smallbreak
\noindent
\textbf{Example 1:} Suppose $N_1 = 2$ and the columns of $\bX$ lie in two spatially separate clusters.
 The data distribution is illustrated in the left plot of Fig. \ref{fig:circlerank}.  The data points lie on two separate arcs of the unit $\ell_2$-norm circle with lengths $\tau_1$ and $\tau_2$. 
We further assume that each of the two arcs does not overlap with the image of the other arc w.r.t. the origin on the unit circle. 
 The number of data points in the first and second clusters are equal to $n_1$ and $n_2$, respectively.
The following two lemmas compare the number of columns sampled randomly by RIS and SRS required to ensure that at least $m$ data points are sampled from each cluster, where $m < \min \{n_1 , n_2\}$.
% the number of sampled data points from each cluster is greater than or equal to $m$ where $m < \min \{n_1 , n_2\}$.

\begin{lemma}
Suppose $N_1 = 2$ and the distribution of the data columns follows the assumptions of Example 1. % as in Fig. \ref{fig:circlerank}.
%The number of data points in the first and second clusters are $n_1$ and $n_2$, respectively.
 \textcolor{black}{ Assume  $\beta \ge 2 + \frac{3}{m} \log \frac{4}{\delta}$.} If the number of columns sampled by RIS with replacement is greater than
$
\beta m \frac{N_2}{\min_i \{ n_i \}},
$
then the number of data points sampled from each data cluster is greater than or equal to $m$, with probability at least $1 -  \delta$.
\label{lm:circle_index}
\end{lemma}

\begin{lemma}
Suppose $N_1 = 2$, the distribution of the data columns follows the assumptions of Example 1 as in Fig. \ref{fig:circlerank}
, $\beta \ge 2 + \frac{3}{m} \log \frac{4
}{\delta},$ and $  \tau_2 + \tau_1  < \pi $. 
%In addition, define $\calS_1$ and $\calS_2$ as the sets of all  points of the unit circle on the first and second arcs, respectively, and assume that the set $\left\{ \by \in \mathbb{S}^{1} \: \Big| \: |\by^T \bz_1| = |\by^T \bz_2| , \bz_1 \in \calS_1 , \bz_2 \in \calS_2  \right\}$ is empty. 
If the number of columns sampled by SRS (with replacement) is greater than
$
\beta m \frac{2 \pi}{\pi - |\tau_2 - \tau_1| } \: , 
$
then the number of data points sampled from each data cluster is greater than or equal to $m$, with probability at least $1 -  \delta$.
\label{lm:circle_spatial}
\end{lemma}

Lemma \ref{lm:circle_index} establishes a sufficient condition on the number of columns sampled by RIS, which is shown to be linear in % that the sufficient number of sampled columns by RIS is linear with 
$\frac{N_2}{\min_i n_i}$ since with RIS the probability of sampling from the $i^{\text{th}}$ cluster is $n_i / N_2$.
Thus, if the populations of the data clusters are unbalanced (i.e., not of the same order), we will need to sample a large number of points to ensure the sampled columns are descriptive, i.e., enough points are drawn from every cluster.
On the other hand, Lemma \ref{lm:circle_spatial} shows that the sufficient number of columns sampled by SRS is independent of the cluster populations since the probabilities of sampling from the first and the second clusters with SRS are equal to $\frac{\pi + \tau_1 - \tau_2}{2 \pi}$ and $\frac{\pi + \tau_2 - \tau_1}{2 \pi}$, respectively. These sampling probabilities are proportional to the surface areas covered by the clusters on the unit sphere (c.f. Definition \ref{def:surf_area}) independently from the cluster populations. % and they are proportional to the area covered by the clusters.

\smallbreak
\noindent
\textbf{Example 2:} In this example,
we consider a different clustering structure in which the columns of the data matrix $\bD$ lie in a union of $s$ linear subspaces  (the subspace clustering structure \cite{vidal2011subspace,rahmani2015innovation}).
Assumption \ref{asum_union}  formalizes the underlying data model.

\begin{assumption}
The columns of $\bD$
lie in a union of $s$ random $(r/s)$-dimensional linear subspaces. i.e.,
$\bD = [\bU_1 \bQ_1 \: ... \: \bU_s \bQ_s]$, where
$\bQ_i \in \mathbb{R}^{ r/s \times n_i}$, $n_i$ is the number of data points in the $i^{\text{th}}$ subspace, and
 $\bU_i \in \mathbb{R}^{N_1 \times r/s}$ is an orthonormal basis for the $i^{\text{th}}$ subspace. The data points in each subspace are distributed uniformly at random, $\sum_{i = 1}^s n_i = N_2$, and $\underset{i}{\min}\: {n_i} \gg r/s$.
\label{asum_union}
\end{assumption}
% \noindent

\noindent
Define $\{p_i\}_{i=1}^s$ as the sampling probabilities from the data clusters. The following lemma provides a sufficient condition on the number of randomly sampled columns to ensure that they span the column space of $\bD$.

\begin{lemma}
Suppose Assumption \ref{asum_union} holds, the rank of $\bD$ is equal to $r$, and RIS or SRS is used to sample $n$ columns (with replacement). If
\begin{eqnarray}
n \ge \frac{1}{\underset{i}{\min} \: p_i} \: \xi_{\max} \left( 2 + \frac{3}{\xi_{\min}} \log\frac{2 \:s}{\delta} \right) , 
\label{eq:suf_cs}
\end{eqnarray}
then the sampled columns span the column space of $\bD$ with probability at least $1 - 2 \delta - 2 \sum_{i = 1}^s n_i^{-3}$, where $
 \xi_{\min} = {10} \: c \: {\max(r/s , \log \underset{i}{\min}\: n_i )} \log \frac{2 r}{\delta}$, $
\xi_{\max} = {10} \: c \: {\max(r/s , \log \underset{i}{\max}\: n_i )} \log \frac{2 r}{\delta}
$, and $c$ is a real constant.

\label{lm:CS_SRS}
\end{lemma}

\noindent
Thus, the sufficient number of sampled columns using RIS and SRS is of order $\calO(\frac{r}{\underset{i}{\min} \: p_i})$. With RIS, $\underset{i}{\max} \: 1/p_i = N_2 / \underset{i}{\min} \: n_i $. Hence, if the data is unbalanced, the sample complexity of RIS is high.
If we use SRS for column sampling,  
the probability of sampling from the $i^{\text{th}}$ cluster is equal to $$\mathbb{P} \left[ \max | \phi^T \bU_i \bQ_i |   > \underset{j \neq i}{\max} \left( \max | \phi^T \bU_j \bQ_j | \right)   \right].$$ 
In contrast to RIS where the sampling probability solely depends on the population ratio, in SRS it also depends on the structure of the data. We show that if the data follows Assumption 1 and the number of data points in each subspace is sufficiently large, the probability of sampling from a data cluster can be independent of the population ratio.  Before stating this result in Lemma \ref{lm:srs_sub}, we define $i^{'} := \argmin n_i $, and define,
\[
p(\epsilon,s) = \mathbb{P} \left[ \| \phi^T \bU_{i^{'}}  \|_2^2 > (1+\epsilon) \: \underset{%i \atop 
i \neq i^{'}}{\max} \| \phi^T \bU_{i}  \|_2^2 \right], 
\]
for some $\epsilon > 0$. If $\epsilon$ is sufficiently small, $p(\epsilon,s)$ is approximately equal to $1/s$. For instance, if $s = 2$, the distribution of $x = \| \phi^T \bU_{1}  \|_2^2 / \| \phi^T \bU_{2}  \|_2^2$ is the \textit{F}-distribution $f(x; r/2,r/2)$ and $f(1; r/2,r/2) = 1/2$ \cite{mood1950introduction}. 

\begin{lemma}
For any $0< \delta <1$, there exits an integer $\hat{n}_\delta$ such that if $n_{i^{'}} > \hat{n}_\delta$, then $p_{i^{'}}> p(\epsilon,s) (1 - \delta)$. 
\label{lm:srs_sub}
\end{lemma}

Per Lemma \ref{lm:srs_sub}, the probability of sampling from a subspace can be independent from the population ratio given there are enough data points in the subspaces, i.e., the probability of sampling from a data cluster can be arbitrarily close to $1/s$ even if ${{\min_i n_i}}/{N_2}$ is close to zero. Note that
this does not mean that each subspace should have a large number of data points for SRS to yield a descriptive data sketch. For instance, suppose $\bD$ follows Assumption 1 with $r = 100$, $s = 50$, $\{ n_i \}_{i = 1}^{25} = 20$, and $\{ n_i \}_{i = 26}^{50} = 500$ (The data lies in the union of 50 2-dimensional subspaces, $\underset{i}{\min} \: n_i = 20$, and $N_2/\underset{i}{\min} \: n_i = 25$).
The right plot of Fig. \ref{fig:circlerank} shows the rank of randomly sampled columns versus the number of sampled columns. RIS samples about 4000 columns to capture the column space versus 200 columns for SRS. % is roughly 200.

%\subsection{SRS balances the data distribution}
\subsection{Balanced sketching}
A marked feature of SRS is that the sampling probabilities depend on the spatial distribution of the data. Therefore, even when the distribution of the data is highly unbalanced, SRS can yield balanced data sketches. % obtained by SRS can be balanced.
For instance, suppose the data follows the distribution shown in the left plot of Fig. \ref{fig:circlerank}
with $\tau_1 = \tau_2$. Since $\tau_1 = \tau_2$, SRS samples a number of points of the same order from each cluster with high probability. Therefore, the data sketch obtained by SRS is balanced even if the given data is not.
This feature is crucial in big data analysis. As an example, consider a scenario where % $n_1$ and $n_2$ are not in the same order. Without loss of generality,
$n_1 \gg n_2$. If some data clustering algorithm is applied to identify two cluster centers, it will select both centers from the first cluster (thus fails to recognize the underlying data structure) as it seeks to minimize the distances between the data points and the cluster centers. % Thus, it fails to recognize the underlying data structure. % The reason is that it tries to minimizes the distance between the data points and the cluster centers. 
However, if the clustering algorithm is applied to a data sketch obtained through SRS, it can identify appropriate cluster centers since SRS balances the distribution of the data in the sketch.

%As another example, suppose the data follows Assumption \ref{asum_union} with $r = 100$, $s = 50$, and $N_1 = 500$, i.e., the data points lie in a union of 50 2-dimensional linear subspaces. $\{ n_i \}_{i=1}^{25} = 10$ and $\{ n_i \}_{i=26}^{50} = 300$.
%We use SRS and RIS to sample 500 columns randomly.
%Fig.  \ref{fig:example_singularvalues} shows the singular values of $\bD$, the data sketch obtained by SRS and the data sketch obtained by RIS. One can observe that the condition number of the matrix obtained by SRS is almost 6 times greater than the condition number of given data because
%it balances the distribution of the data. One can observe that the singular values of the matrix obtained by RIS drops quickly and it is clear that RIS does not provide a descriptive data sketch with 500 samples.

\subsection{Computational complexity}
%Matrix $\bQ$ is the projection of the columns of $\bX$ into a random $n$-dimensional subspace. In the last step of the algorithm, we can form the matrix $\bY$ using the columns of $\bQ$. Thus, we can reduce the dimensionality of the data on both column and row sides, i.e., the size of output data sketch would be $n \times n$.
%
The computational complexity of SRS can be reduced by applying Algorithm 1 to a sketch of the rows of $\bD$.  Define matrix $\bD^{'} \in \mathbf{R}^{p \times N_1}$ as $\bD^{'} = \bS \bD$. We consider three choices for the matrix $\bS$. One choice is where the rows of $\bS$ are a random subset of the standard basis, which amounts to random row sampling. However, a sufficient number of rows should be sampled to ensure that the underlying structures are preserved. If the sampling algorithm is applied to $\bD^{'}$ obtained by row sampling, the complexity is reduced from $O(n N_1 N_2)$ to $O(n p N_2)$. 
The second choice is to select $\bS$ from \emph{sparse} random embedding matrices \cite{kane2014sparser,dasgupta2010sparse}. Since a sparse random embedding matrix contains many zeros, a much reduced number of multiplications is needed to perform the random embedding step akin to random row sampling.
% Similar to row sampling matrix, the sparse random embedding matrix contains many zeros. Thus, much less number of multiplications is required to perform the random embedding step.
The third choice is to use a matrix $\bS$ with entries that are independent binary random variables ($\pm 1$ with equal probability). Random embedding using such a matrix does not involve any numerical multiplications and was  shown to yield embedding performance that closely approaches that of random Gaussian matrices \cite{achlioptas2003database}.
Algorithm 1 can be applied to $\bD^{'}$ since the embedding matrix $\bS$ preserves the essential information in $\bD$.
% Since the embedding matrix $\bS$ preserves the essential information, Algorithm 1 can be applied to $\bD^{'}$.

\begin{figure*}[t!]
    \includegraphics[width=0.9\textwidth]{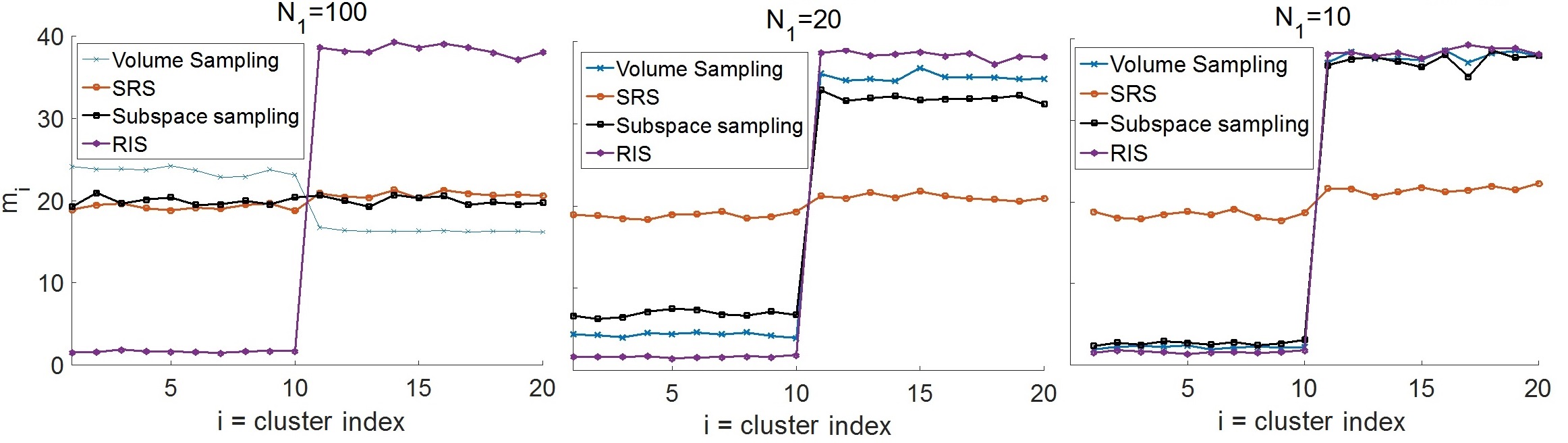}
    \centering
     \vspace{-0.1in}
    \caption{The number of sampled columns from each data cluster for different values of $N_1$. }
    \label{fig:example1}
%   \vspace{-.5cm}
%\label{fig:circlerank}
\end{figure*}

\subsection{Discussion and conclusion}
This paper proposed a new random data sketching tool, which carries out the random sampling in the spatial domain. We emphasize that SRS is not meant to be a replacement for, nor a modification to, RIS. Rather, it is a new sketching tool which has its own applicability. We showed that SRS can provide a balanced data sketch and has higher chance of sampling from rare events compared to sampling with RIS. 
Unlike matrix approximation based methods which require the cluster centers to be independent, linear independence is not important for SRS. %This feature makes SRS appealing for big data applications because the sampling algorithm can be applied to  a sketch of data obtained by random embedding of the data points into a lower dimensional space. 

\section{Numerical Experiments}
In this section, we present a set of numerical experiments using both synthetic and real data to showcase the effectiveness of SRS in preserving the underlying data structure. 
% that although SRS is quite simple, it effectively preserves the structure of data.

\subsection{Balanced data sketching}
Suppose the data follows the subspace clustered structure in Assumption \ref{asum_union} with $r = 40$, $s  =20$, $\{ n_i \}_{i=1}^{10} = 30$, and $\{ n_i \}_{i=11}^{20} = 700$. Thus, the columns of $\bD$ lie in a union of 20 2-dimensional linear subspaces and the distribution of the data is quite unbalanced. The distribution of the data points in the sketch obtained by SRS is compared to that obtained by RIS and by two other adaptive column sampling methods, namely, subspace sampling \cite{drineas2006subspace} and volume sampling \cite{deshpande2010efficient,deshpande2006matrix}. We sample a total of 400 columns. For the subspace sampling method, we use $\min (r , N_1)$ right singular vectors to compute the sampling probabilities.
Volume sampling is an iterative column sampling method that samples one column at a time. It projects the data on the complement space of the sampled data points, thus stops sampling after roughly $\hat{r}$ steps, where $\hat{r}$ is the rank of data. In this experiment, we apply volume sampling multiple times to sample 400 data columns (in each time the sampled columns are removed from the data).

Fig. \ref{fig:example1} shows the number of data points $m_i$ sampled from each data cluster as a function of the cluster index $i = 1,\ldots, 20$.  The plots are obtained by averaging over 100 independent runs.
% and $m_i$ is the number of sampled columns from $i^{\text{th}}$ cluster.
If  $N_1 > 40$, the subspaces are independent with high probability. Clearly, when $N_1 = 100 > 40$, almost all the sampling algorithms can yield a balanced data sketch except for RIS. However, as $N_1$ decreases (e.g., $N_1 = 10$ and $N_1 = 20$), the subspaces are no longer independent. In this case, only SRS is shown to yield a balanced data sketch. This is due to the fact that the sampling probability with RIS depends on the sizes of the populations of the clusters, and adaptive sampling only guarantees that the span of the sampled columns well approximates the column space of the data.

\subsection{Spatially fair random sampling}
\begin{figure*}[t!]
    \includegraphics[width=0.9\textwidth]{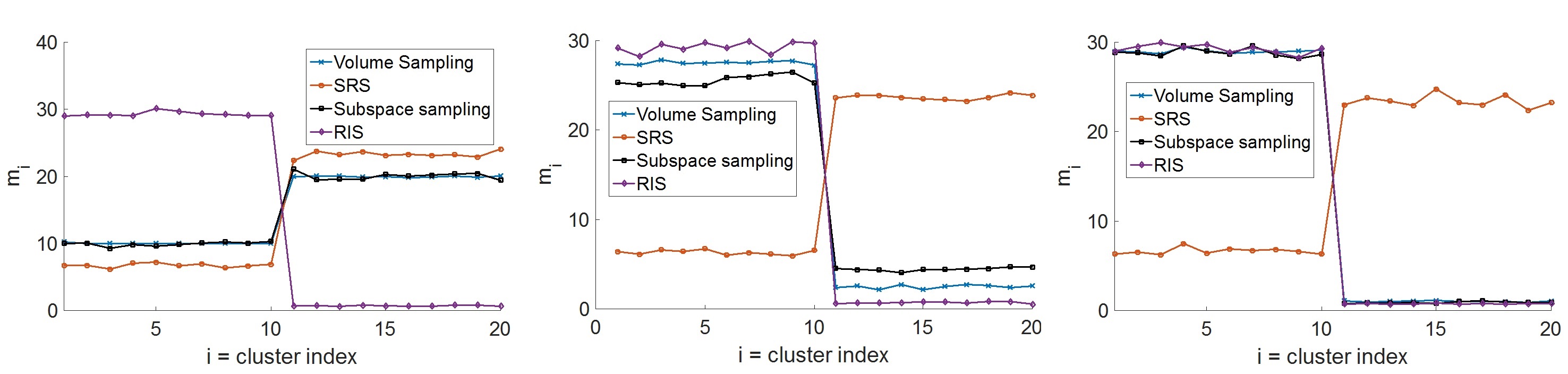}
    \centering
     \vspace{-0.1in}
    \caption{The number of columns sampled from each data cluster. In the left plot, the sampling algorithms are applied directly to the data. In the middle and right plot, the sampling algorithms are applied to a sketch of the data obtained through random embedding.}
    \label{fig:example122}
%   \vspace{-.5cm}
%\label{fig:circlerank}
\end{figure*}

Similar to the previous subsection, assume the data lie in a union of 20 linear subspaces $\{ \calS \}_{i=1}^{20}$. The dimension of subspaces $\{ \calS \}_{i=1}^{10}$ is equal to 2 and the dimension of subspaces $\{ \calS \}_{i=11}^{20}$ is equal to 4. The number of data points lying in each of the subspaces  $\{ \calS \}_{i=1}^{10}$ is equal to 3200, while the number of data points lying in each of the subspaces $\{ \calS \}_{i=11}^{20}$ is only 80.
Thus, $N_2 = \left(10\times 80 + 10\times 3200 \right)$. Importantly, the subspaces with the lower dimension contain 40 times more data points than those with the higher dimension. Since more data points are required to represent clusters with higher dimensions, we naturally desire that the sampling algorithm samples more points from such clusters. % with higher dimensions since more data points are required to represent such clusters. 
In this experiment, 300 data columns are sampled. Fig. \ref{fig:example122} shows the average number of data points sampled from each cluster.
In the left plot, $\bD \in \mathbb{R}^{100 \times N_2}$.  Thus, the subspaces are independent with high probability. One can observe that all the sampling algorithms except RIS follow a similar pattern of sampling, namely, sample more data points from the 10 subspaces with the higher dimensions. 

In the middle and right plots, the sampling algorithms are applied to a sketch of the data obtained by randomly embedding the data into a lower dimension space. The sampling algorithms are applied to $\bD^{'} = \bS \bD$, where in the middle plot $\bS \in \mathbb{R}^{20 \times N_1}$ and in the right plot $\bS \in \mathbb{R}^{10 \times N_1}$ and $N_1 = 100$. The matrix $\bS$ is a random binary matrix whose elements are independent random variables with values $\pm 1$ with equal probability. % The elements of $\bS$ are independent random variables. % Each element is equal to $\pm 1$ with equal probability.
One can observe that only SRS exhibits the same sampling pattern of the left plot. Since the matrix approximation based methods seek a good approximation of the dominant left singular vectors of the data, they cannot recognize the underlying clustering structure of the data if the clusters are linearly dependent. 

%One can observe that it is only the SRS method can provide expressive and also fair data sketch with all values of $N_1$. When $N_1$ is small, the subspaces become linearly dependent and it makes the matrix approximation methods such as the volume sampling and subspace sampling unable to understand the structure of the data. 
%

%
\subsection{Column sampling for classification}
We test the proposed approach with real data, the MNIST database \cite{mnist2010}.
The data consists of $28 \times 28$  handwritten digit images. The MNIST database contains \textcolor{black}{50000} and 10000 images for training and testing, respectively, and 10000 images  of validation data.
In this experiment, we consider a binary classification problem.
The first class corresponds to numbers between 0 to 4, and the second class corresponds to numbers greater than or equal to 5.
The training data $\textbf{Tr}_1$ corresponding to the first class is constructed as $\textbf{Tr}_1 = [\bD_0 \: \: \bD_1 \: \: \bD_2 \: \: \bD_3 \: \: \bD_4]$, where $\bD_0 \in \mathbb{R}^{784 \times 4900}$, $\bD_1 \in \mathbb{R}^{784 \times k}$, $\bD_2 \in \mathbb{R}^{784 \times 4900}$, $\bD_3 \in \mathbb{R}^{784 \times k}$, $\bD_4 \in \mathbb{R}^{784 \times 4900}$ ($k$ is a changing parameter as shown in Table \ref{tabl 1}). The columns of $\bD_0$, $\bD_1$, $\bD_2$, $\bD_3$, and $\bD_4$, are randomly sampled training images corresponding to digits 0, 1, 2, 3, and 4, respectively.
Similarly, the training data corresponding to the second class is $\textbf{Tr}_2 = [\bD_5 \: \: \bD_6 \: \: \bD_7 \: \: \bD_8 \: \: \bD_9]$, where $\bD_5 \in \mathbb{R}^{784 \times 4900}$, $\bD_6 \in \mathbb{R}^{784 \times k}$, $\bD_7 \in \mathbb{R}^{784 \times 4900}$, $\bD_8 \in \mathbb{R}^{784 \times k}$, $\bD_9 \in \mathbb{R}^{784 \times 4900}$.
The columns of $\bD_5$, $\bD_6$, $\bD_7$, $\bD_8$, and $\bD_9$, are randomly sampled training images corresponding to digits 5, 6, 7, 8, and 9, respectively.
Thus, the entire training data is $\textbf{Tr} = [\textbf{Tr}_1 \: \: \textbf{Tr}_2]$. 

We do not use all the columns of $\textbf{Tr}$ to train the classifier, rather we sample 1000 columns randomly from each class (2000 in total) and use these sampled columns to train the classifier.
The classifier is a two-layer fully connected neural network with 400 neurons in each layer.
Table 1 compares the classification accuracy for different values of $k$.  When $k$ is small, the distribution of the data is unbalanced across classes.
As shown the performance gap of RIS relative to SRS increases as $k$ decreases. For instance, when $k = 300$, the classification accuracy achieved based on SRS is substantially higher by about 5 percent. % higher, which is a substantial difference.

\begin{table}
\centering
\caption{\textcolor{black}{Classification accuracy of the classifier}}
\begin{tabular}{| c | c | c | c | c |}
\hline
Sampling method / $k$ & 4900 & 2000 & 1000 & 300 \\
 \hline
SRS & 0.9671 & 0.9615 & 0.9587 & 0.9436 \\
 \hline
RIS & 0.9615 & 0.9561 & 0.9443 & 0.8968 \\
 \hline
\end{tabular}
\label{tabl 1}
\end{table}

%random:
%0.9611 , 0.9567 [.9448 .9462 .4998 .9443]        [.8991 .893   0.8958 .8962]

\newpage
\bibliographystyle{IEEEtran}
\bibliography{IEEEabrv,bibfile}

\end{document}